\documentclass[conference]{IEEEtran}
\IEEEoverridecommandlockouts
\usepackage{cite}
\usepackage{amsmath,amssymb,amsfonts}
\usepackage{algorithmic}
\usepackage{graphicx}
\usepackage{subfig}
\usepackage{textcomp}
\usepackage{xcolor}
\usepackage{mathtools}
\usepackage{booktabs} 
\usepackage{fancyhdr}
\usepackage[]{mdframed}
\usepackage{url}
\def\BibTeX{{\rm B\kern-.05em{\sc i\kern-.025em b}\kern-.08em
    T\kern-.1667em\lower.7ex\hbox{E}\kern-.125emX}}

\fancypagestyle{specialfooter}{%
  \fancyhf{}
  
  \fancyhead[L]{2024 the 6th World Symposium on Artificial Intelligence (WSAI 2024)}
}

\begin{document}

\title{Venn Diagram Prompting : Accelerating Comprehension with Scaffolding Effect\\
}

\author{\IEEEauthorblockN{Sakshi Mahendru}
\IEEEauthorblockA{
\textit{Palo Alto Networks}\\
California, USA \\
}
~\\
\and
\IEEEauthorblockN{Tejul Pandit}
\IEEEauthorblockA{
\textit{Palo Alto Networks}\\
California, USA \\
}


}

\maketitle

\begin{abstract}
We introduce Venn Diagram (VD) Prompting, an innovative  prompting technique which allows Large Language Models (LLMs) to combine and synthesize information across complex, diverse and long-context documents in knowledge-intensive question-answering tasks. Generating answers from multiple documents involves numerous steps to extract relevant and unique information and amalgamate it into a cohesive response. To improve the quality of the final answer, multiple LLM calls or pretrained models are used to perform different tasks such as summarization, reorganization and customization. The approach covered in the paper focuses on replacing the multi-step strategy via a single LLM call using VD prompting. Our proposed technique also aims to eliminate the inherent position bias in the LLMs, enhancing consistency in answers by removing sensitivity to the sequence of input information. It overcomes the challenge of inconsistency traditionally associated with varying input sequences. We also explore the practical applications of the VD prompt based on our examination of the prompt's outcomes. In the experiments performed on four public benchmark question-answering datasets, VD prompting continually matches or surpasses the performance of a meticulously crafted instruction prompt which adheres to optimal guidelines and practices.
\end{abstract}


\section{Introduction} \label{Intro}
\thispagestyle{specialfooter}
Retriever-based approaches using LLM-based generators (RAG) have become increasingly popular for knowledge-intensive NLP tasks. The advent of Large Language Models (LLMs) like GPT-4 has been instrumental in significantly improving RAG systems \cite{gao2024retrievalaugmented}. However, LLMs have certain shortcomings, such as providing out-of-date information, false reasoning, or factually incorrect responses. The language models' performance also degrades significantly when changing the position of relevant information, indicating that models struggle to robustly access and use information in long input contexts. RAG partially addresses some of these concerns and improves the accuracy of responses by building knowledge into the LLM from the retrieval stage \cite{rag1}. However, the underlying nature of LLMs makes the generation module of RAG systems lossy \cite{rag2}, which impacts the entire process. With increasing context length and diverse information present in the retrieved documents, the generator module struggles to comprehend the complete knowledge context and sometimes produce inconsistent results, especially if there is unordered, repetitive, or redundant information across the documents \cite{gao2024retrievalaugmented, li2022survey}. To sidestep these issues, existing systems include multiple steps or devise a complex logic to achieve factually correct results.

With increasing context length for LLMs, it is able process more documents than before, however it also exacerbates its existing issues such as position bias, hallucinations, erroneous reasoning and inconsistent responses. Our proposed innovative prompting technique, VD prompting significantly mitigates these issues. It performs all the expected steps for answer generation in a single shot with efficacy. This is made possible as the prompt is guided using the Venn diagram approach to first organize the information from the provided context before generating an answer. The organization of information in a structured manner makes it easier to identify overlapping elements and thereby act as a scaffolding that improves the comprehension, leading to efficient responses. VD prompt also has the unique capability to provide citations along with the relevance of the document to the query, which helps to trace the origin of information. Furthermore, this novel methodology is aimed at mitigating position bias and associated degradation of performance of language models. This mechanism identifies and prioritizes relevant information within the context, allowing models to focus on crucial segments regardless of their position.

Through empirical evaluation, we demonstrate that VD prompting significantly improves the model robustness and performance across various datasets. Our approach not only addresses the aforementioned discussed problems but also enhances the overall utility and applicability of long-context language models in real-world scenarios.

\begin{figure}[htbp]
\centerline{\includegraphics[width=8cm]{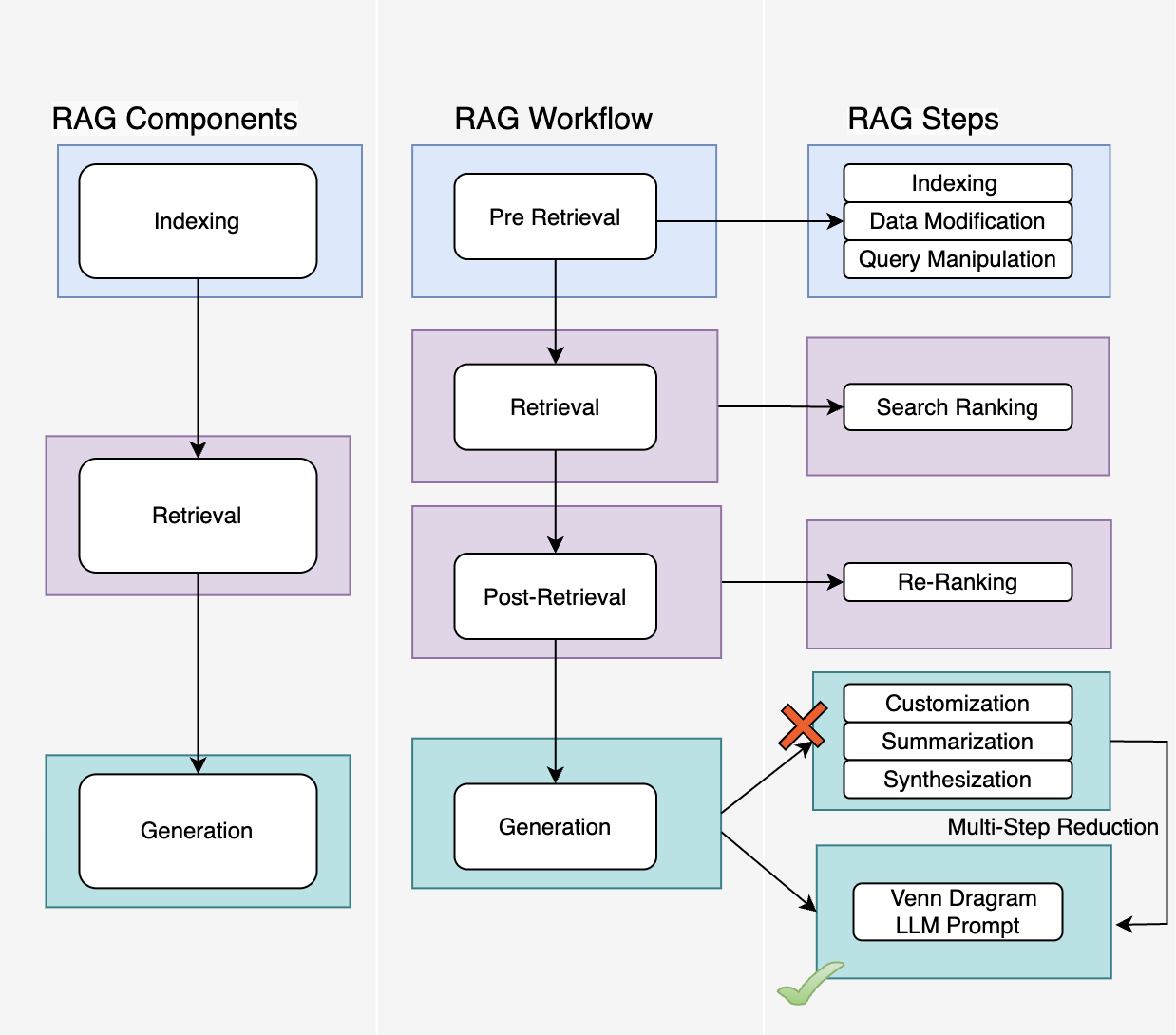}}
\caption{A unified RAG framework\cite{li2022survey} with basic workflow and VD prompt proposed workflow which modifies the generation component. The new workflow presents an innovative approach that condenses the multiple steps of the original workflow via Venn diagram prompting.}
\label{ragChanges}
\end{figure}

Fig. \ref{ragChanges} illustrates the differences in the original workflow and VD prompt modified proposed workflow in RAG framework. The generation module navigates through diverse information from retrieved data sources to create a coherent and relevant answer. In the original workflow, this could involve filtering, synthesis, summarization, customization, and other methods to improve the quality of an answer. Our proposed workflow suggests a promising prompting technique, VD prompting which can perform all these steps together and present an accurate and coherent answer.

\begin{figure}[htbp]
\centerline{\includegraphics[width=10cm]{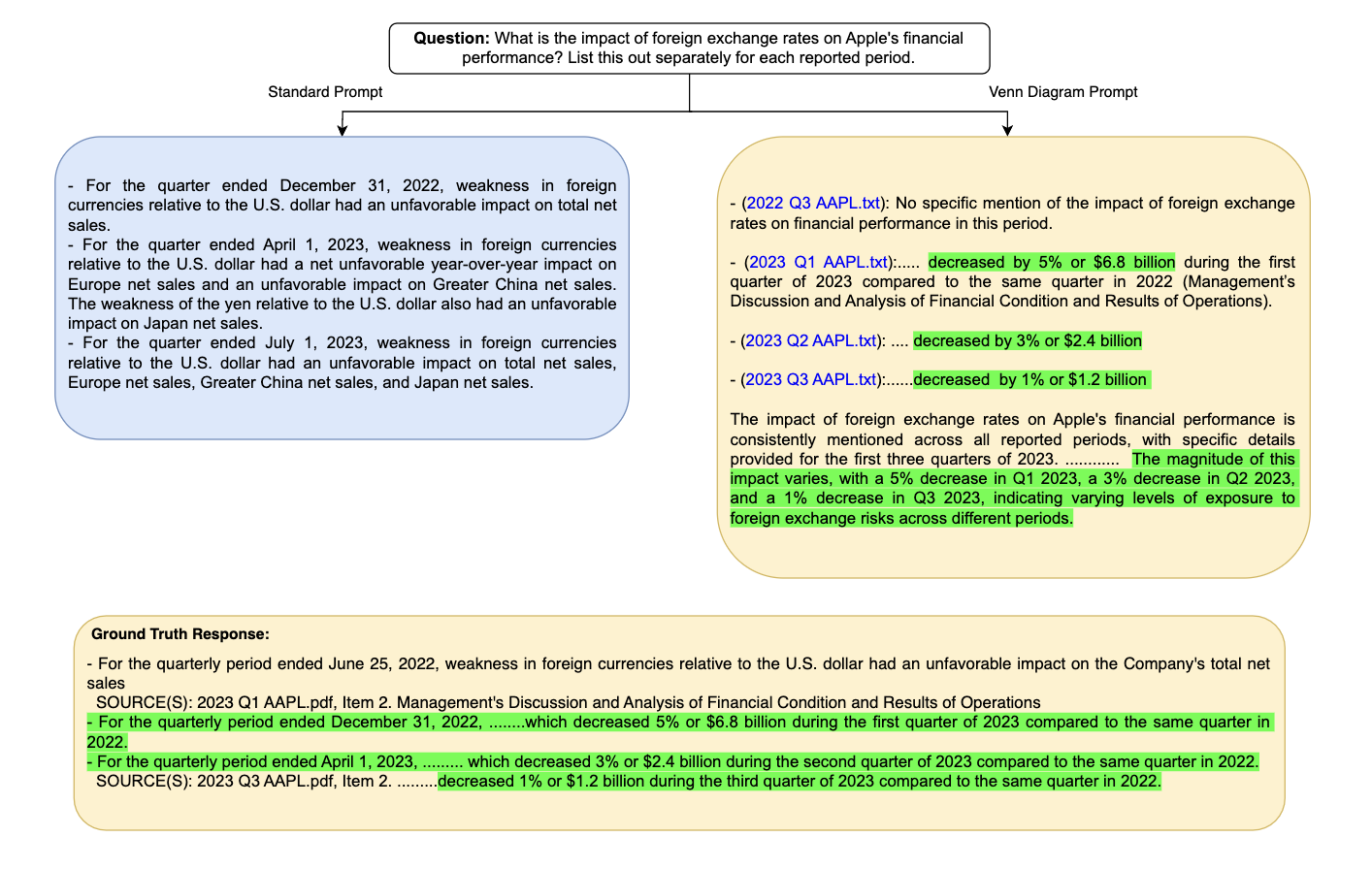}}
\caption{Comparison between standard and VD prompt for the same query and comparison with Ground Truth Response}
\label{ansCompare}
\end{figure}

As observed in Fig. \ref{ansCompare}, when we provide the same query to an instruction-based prompt and a VD prompt with the same document(s) context for reference, VD prompt extracts the most relevant details from the query, aligning closely with the expected outcomes. Standard prompt, on the other hand, does extract relevant details but fails to capture complete information. This is validated by comparing the two responses with the Ground Truth Response provided for the query. There is a significant overlap in the specific details between the Ground Truth Response and the response generated from the VD prompt.

We delve into further details by discussing Related Work in Section \ref{relatedWork} followed by articulating the exact approach taken with respect to the VD prompt in Section \ref{approach} and cover the experimental setup and their results in Sections \ref{expSetup} and \ref{evaluationResults} respectively. Sections \ref{adv} provides advantages that we foresee by using the VD prompt and Section \ref{futureScope} provides the future scope and additional use-cases that we envision with our approach followed by concluding our findings in Section \ref{conclusion}.

\section{Related Work} \label{relatedWork}

\subsection{Prior Findings}

There has been a significant boom in recent years with prompt engineering techniques being developed to uplift LLM performance for specific response generation tasks.
Widely popular prompting techniqies such as Chain-of-thought\cite{wei2023chainofthought}, Reflexion\cite{shinn2023reflexion} and Directional Stimulus \cite{li2023guiding} have unleashed the LLM's capabilities to arrive at factually correct, accurate, less biased and non-hallucinated responses. It is needed as model scaling and fine tuning face challenges with latest information and reasoning capabilities. Moreover, the process of fine-tuning an LLM typically encounters difficulties in generating large quality-rich datasets for specific tasks or issues with models being under-fitted or over-fitted due to the requirement of handling multiple hyperparameters. LLMs are a black box with no visibility on how they are trained and this shortcoming is reflected in the fine-tuning process as well. Therefore, prompting techniques such as  \cite{wei2023chainofthought, shinn2023reflexion, li2023guiding} are cost-effective and come in handy. They help break down complex tasks into easier intermediate steps enabling the LLMs to improve the performance. 

The latest framework, RAG \cite{li2022survey} presents yet another cost-effective alternative than fine-tuning LLMs as continuous knowledge update is a major challenge for fine tuned models. In RAG systems, the efficacy of the final output results is dominated by the quality of the generator module. Consequently, the effectiveness of the entire RAG system is bounded by the capabilities of its generator, which is intricate as diverse information sources need to be integrated to provide a holistic viewpoint \cite{li2022survey}.

The existing work \cite{liu2023lost, ren2024identifying} discusses LLMs challenges related to position bias, i.e. the LLMs encounter inconsistency in answers due to the position of relevant text. There is notable degradation in performance when relevant information is repositioned, suggesting that existing language models struggle to consistently utilize information in lengthy input contexts.

\subsection{Challenges}
This section focuses on the challenges of the RAG text generation module. Collating information from retrieved documents presents significant challenges due to the inherent complexity, diversity, and lengthiness of the documents\cite{zhao2024retrievalaugmented}. It involves multiple steps like information extraction, deduplication, summarization, and synthesis of only the relevant information with reasoning to formulate an accurate answer. The advanced RAG pipelines implement this multi-step approach using LLMs to generate the answer\cite{li2022survey}, which brings variance and bias due to multiple LLMs involved. It also adds to the maintenance cost of the pipeline. Additionally, the final generated response lacks indication of the source for a specific detail. Though the information is included, it lacks proper citation of sources, posing a challenge for readers who wish to delve deeper into a specific detail and ascertain its origin. 

Crafting a prompt for the answer generation module can be intricate, typically involving several phases. When generating response from multiple retrieved documents, two scenarios arise: a) A document may be pertinent but lacks the answer. b) Multiple documents are needed to synthesize the answer. Thus, ensuring the generation of precise and high-quality answers necessitates amalgamating the documents, sieving out irrelevant details, and forming segments that encapsulate unique and overlapping information relevant to the given query.

\section{Approach} \label{approach}
We formulate the prompt by interpreting the query provided as the Universal Set ($\xi$) and all the documents/chunks used as reference interpreted as unique sets, each represented by its own individual circle. A combination of all these unique sets, denoted by ($A$, $B$, $C$,...) are such that they are either a subset of the $\xi$, partially overlap over $\xi$ or lie completely outside of the $\xi$, denoted by $\xi\textsuperscript{c}$. The core ideology behind the prompt is to ensure that we extract all the essential details pertaining to the given query while simultaneously rejecting any information that lies in the $\xi\textsuperscript{c}$. The expectation lies in leveraging the concept of Venn diagrams and set theory while merging information from a single/multiple document/chunk(s). We represent the final response generated by VD prompt with $\xi\textsuperscript{'}$, constrained by the provided sources.
 Additionally, the prompt structuring adds to building the knowledge of organizing pertinent information and adding structure, thus providing a more consistent response. Based on our observation,  running the same prompt multiple times does not  change the fact that, if set $A$ $\subset$ $\xi$ in one instance, then it will remain congruous with every subsequent run. Similarly, if a set $D$ does not belong to $\xi$, it will be construed as D $\cap$ $\xi$  = $\O$, where $\O$ indicates a null or empty set, consistently. However, the same set of standards cannot be established for an instruction-based prompting technique due to position bias issue \cite{liu2023lost, ren2024identifying}. We support our claim that VD prompting results are consistent across multiple runs with a small experiment later in this section.

We elaborate our idea with the help of an example. Consider a query - “How to boil eggs?” and there are 6 documents provided as prompt context as shown below. Two sets of prompts are created, one based on specific instructions which adheres to optimal guidelines and practices and the other using our proposed VD approach.

\begin{mdframed}
\textbf{Document 1:}

\noindent To boil eggs, simply place them in a pot, cover with water, and bring to a boil. Boiling eggs is a simple task with deep roots in world history. Ancient Egyptians boiled eggs for religious rituals, while WWI soldiers relied on them for sustenance. The humble boiled egg has been a staple across cultures and eras, reflecting humanity's ingenuity and adaptability.

\noindent\textbf{Document 2:}

\noindent To boil eggs perfectly, gently place them in a pot and cover with cold water by about an inch. Bring the water to a rolling boil over high heat, then immediately remove the pot from the heat and cover. Let the eggs sit in the hot water for 6-12 minutes, depending on how you like your eggs cooked. Eggs are a powerhouse of nutrition for athletes. Rich in high-quality protein, they aid muscle growth and repair. Beyond protein, eggs provide essential vitamins and minerals that support energy, red blood cell formation, and bone health. Their versatility makes them a convenient sports nutrition choice.

\noindent\textbf{Document 3:}

\noindent To boil hard-boiled eggs, place the eggs in a single layer in a pot and cover with cold water by 1 inch. Bring to a boil over high heat, then remove from heat, cover, and let sit for 12 minutes for hard-boiled. For soft-boiled eggs, boil for 6 minutes, and for medium-boiled, 8 minutes. Eggs can be enjoyed on their own, made into deviled eggs, used in egg salad, or incorporated into dishes like quiche, frittatas, and breakfast sandwiches. Eggs are a versatile and nutritious food, produced by various birds. The yolk contains most of the egg's nutrients, while the white is primarily protein and water. Eggs can be prepared in countless ways and are used in cuisines worldwide. They offer potential health benefits beyond their culinary uses.

\noindent\textbf{Document 4:}

\noindent Eggs are a nutrition powerhouse, benefiting athletes and fashion enthusiasts alike. The high protein content supports muscle development, while vitamins and minerals promote skin health and immune function. The convenience of eggs makes them a practical choice for active lifestyles.

\noindent\textbf{Document 5:}

\noindent Fashion has long reflected and influenced world history. The French Revolution led to more accessible styles, while wars brought about practical, functional clothing. Today, fashion continues to express identity, values, and aspirations, mirroring the dynamic currents of global events.

\noindent\textbf{Document 6:}

\noindent Sports have mirrored the social, political, and cultural forces shaping world history. Major events, like the Olympic Games, have often reflected geopolitical tensions. The globalization of sports has transformed how people engage with athletic achievements, raising questions about the role of sports in world history.
\end{mdframed}

The standard prompt is designed by providing specific instructions to generate a final answer based on the query provided using the given context and cite the sources wherever applicable. The result obtained for our specific case is illustrated in Fig. \ref{fig:std1}

The VD prompt is instructed to find overlapping information and unique information, relevant to the query before providing a final answer. The query is represented by $\xi$ and all the documents provided as context to the prompt individually act as unique sets. Each document may contain information that either overlaps or diverges from others, and its relevance to the given query may vary. The steps involved prior to answer generation by VD prompt are covered in Fig. (\ref{fig:vda}, \ref{fig:vdb}, \ref{fig:vdc}, \ref{fig:vdd}, and \ref{fig:vde}). Final answer for VD prompt is showcased in Fig. \ref{fig:vd1}.  

\begin{figure}[htbp]
\centerline{\includegraphics[width=\linewidth]{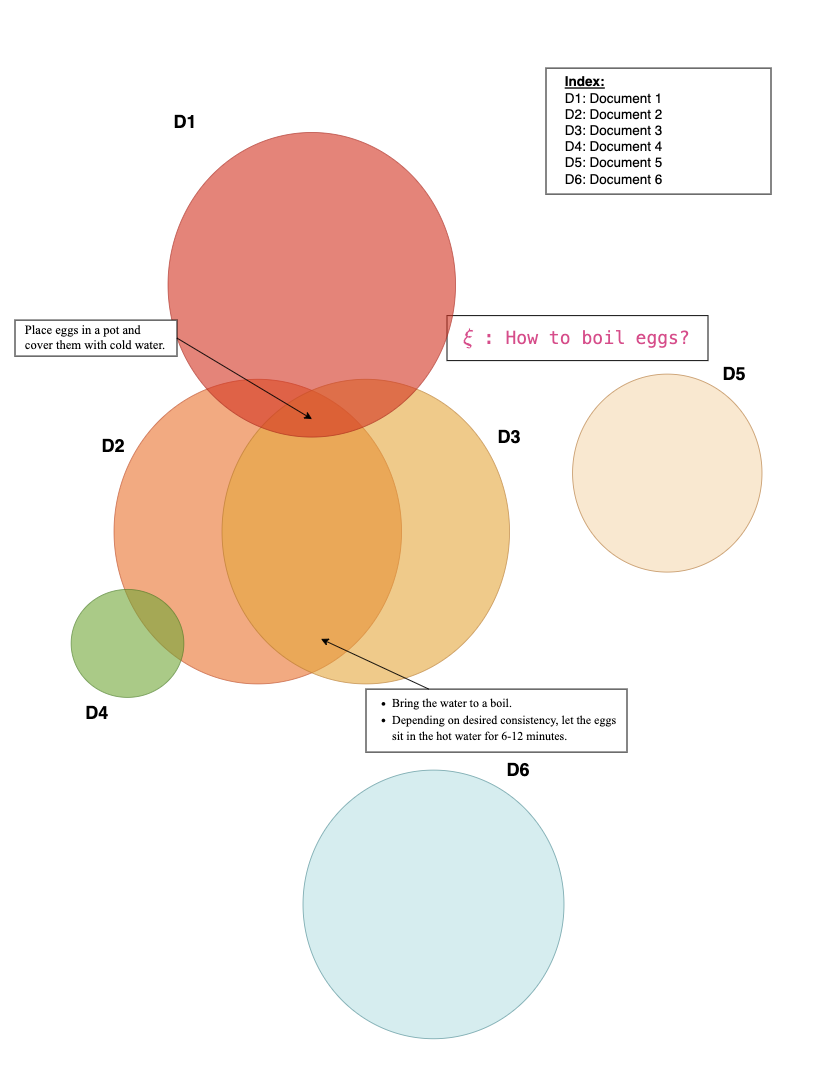}}
\caption{Representation of VD prompt example using Venn Diagram}
\label{fig:vdsets}
\end{figure}

Fig. \ref{fig:vda} solely focuses on listing out all the document sources that are provided as context to identify the unique sets that are present. Overlapping information that is primarily relevant to the query is identified as the next step as shown in Fig. \ref{fig:vdb}. The unique pieces of information present in each document are extracted as illustrated in Fig. \ref{fig:vdc}. Fig. \ref{fig:vdd} and Fig. \ref{fig:vde} indicate the responsibility of the prompt of explaining the document sources derived in the previous 2 steps. A combination of all these steps is responsible for generating the final answer. An illustrative figure explaining with the context of Venn Diagram and the rationale taken behind generating the response is elucidated using Fig. \ref{fig:vdsets}. It also helps to highlight the set notation derived for generating the final answer of our example using the VD prompt approach, which is given by :

\begin{equation}
    \xi\textsuperscript{'} = (D2 \cap D3) \cup (D1 \cap D2 \cap D3)
\end{equation}

To address the claims made towards the beginning of the current section regarding variability introduced in standard prompt response by changing the position of the document and VD prompt's insensitivity to document ordering, we conducted an experiment by implementing a cyclic reordering procedure for documents. In each pass, we assign new identifiers to the documents according to a specific pattern. Since we know that the first 3 documents are most relevant to the query provided, we move the top document towards the bottom of the context in the first pass, i.e. Document 1 is relabeled as Document 6, Document 2 as Document 1, and so on for all other documents.  In subsequent pass, we again update the document identifiers based on the reordered sequence from the previous iteration.  This process ensures that the documents are repositioned. We compare the responses from standard prompt and VD prompt for both passes. The responses generated in both the passes are as captured in Fig. \ref{fig:ex2}. Responses from standard prompt as seen in Fig. \ref{fig:sd2} and Fig. \ref{fig:sd3} affirms the claim. As the relevant document moves from first position to last position, it is not referenced anymore. In contrast, responses from VD prompt as shown in Fig. \ref{fig:vd2} and \ref{fig:vd3} are consistent and support our claim that the position of the document has no impact on the generated final response.

\begin{figure}%
\centering
\subfloat[]{\begin{minipage}{\linewidth}
  \begin{mdframed}[roundcorner=10pt]
    To boil eggs perfectly, gently place them in a pot and cover with cold water by about an inch. Bring the water to a rolling boil over high heat, then immediately remove the pot from the heat and cover. Let the eggs sit in the hot water for 6-12 minutes, depending on how you like your eggs cooked. (Document 2)

  \end{mdframed}
\end{minipage} \label{fig:std1}}
\qquad
\subfloat[]{\begin{minipage}{\linewidth}
  \begin{mdframed}[roundcorner=10pt]
    To boil eggs, as per Document 2 and Document 3, you should start by gently placing the eggs in a pot and covering them with cold water by about an inch. Then, bring the water to a rolling boil over high heat. Once boiling, immediately remove the pot from the heat and cover it. Let the eggs sit in the hot water for 6-12 minutes, depending on your preference for how you like your eggs cooked, with 6 minutes for soft-boiled, 8 minutes for medium-boiled, and 12 minutes for hard-boiled eggs. Document 1 also mentions boiling eggs but does not provide specific instructions like Documents 2 and 3.

  \end{mdframed}
\end{minipage} \label{fig:vd1}}%
\caption{(a) response generated using standard prompt, (b) response generated using VD prompt}%
\label{fig:ex1}%
 \end{figure}

\begin{figure}

\subfloat[]{\begin{minipage}{.45\linewidth}
  \begin{mdframed}[roundcorner=10pt]
    \textit{\textbf{1. Identify the sources}}:

- Circle 1: Document 1

- Circle 2: Document 2

- Circle 3: Document 3

- Circle 4: Document 4

- Circle 5: Document 5

- Circle 6: Document 6

  \end{mdframed}
\end{minipage} \label{fig:vda}}%
\subfloat[]{\begin{minipage}{.5\linewidth}
  \begin{mdframed}[roundcorner=10pt]
    \textit{\textbf{2. Find the Overlapping Areas with respect to query}}:

- Overlapping information between Documents 1, 2, and 3: All provide instructions on how to boil eggs.

  \end{mdframed}
\end{minipage} \label{fig:vdb}}
\qquad
\subfloat[]{\begin{minipage}{\linewidth}
  \begin{mdframed}[roundcorner=10pt]
    \textit{\textbf{3. Find the Distinct Areas with respect to query}}:

- Document 1: Historical and cultural significance of boiled eggs.

- Document 2: Nutritional benefits of eggs for athletes and specific boiling instructions for perfect eggs.

- Document 3: Detailed instructions for boiling eggs to different levels of doneness and the versatility of eggs in cooking.

- Document 4: Nutritional benefits of eggs for athletes and fashion enthusiasts.

- Document 5 \& 6: Do not contain relevant information with respect to the query.
  \end{mdframed}
\end{minipage} \label{fig:vdc}}%
\qquad
\subfloat[]{\begin{minipage}{\linewidth}
  \begin{mdframed}[roundcorner=10pt]
    \textit{\textbf{4. Explain the Overlaps with respect to query}}:

- Documents 1, 2, and 3 all provide methods for boiling eggs, making them directly relevant to the query. They overlap in the basic instruction of boiling eggs but differ in the level of detail and additional context they provide.

  \end{mdframed}
\end{minipage} \label{fig:vdd}}
\qquad
\subfloat[]{\begin{minipage}{\linewidth}
  \begin{mdframed}[roundcorner=10pt]
    \textit{\textbf{5. Explain the Unique Information with respect to query}}:

- Document 1 uniquely mentions the historical and cultural significance of boiling eggs.

- Document 2 uniquely focuses on the nutritional benefits of eggs for athletes and provides a specific method for perfectly boiling eggs.

- Document 3 uniquely provides detailed instructions for achieving different levels of doneness when boiling eggs and discusses the versatility of eggs in cooking.

- Document 4 uniquely mentions the benefits of eggs for both athletes and fashion enthusiasts without providing specific boiling instructions.

  \end{mdframed}
\end{minipage} \label{fig:vde}}%
\caption{Description of each step taken by VD prompt (a) identification of source(s) provided, (b) find overlapping areas w.r.t. query, (c) find unique areas w.r.t. query, (d) explanation for overlapping areas (e) explanation for unique areas}%
\label{fig:ex1-explain}%
\end{figure}


\begin{figure}%
\centering
\subfloat[]{\begin{minipage}{\linewidth}
  \begin{mdframed}[roundcorner=10pt]
    To boil eggs perfectly, gently place them in a pot and cover with cold water by about an inch. Bring the water to a rolling boil over high heat, then immediately remove the pot from the heat and cover. Let the eggs sit in the hot water for 6-12 minutes, depending on how you like your eggs cooked. (Document 1)

  \end{mdframed}
\end{minipage} \label{fig:sd2}}%
\qquad
\subfloat[]{\begin{minipage}{\linewidth}
  \begin{mdframed}[roundcorner=10pt]
    To boil eggs perfectly, you should start by placing the eggs in a pot and covering them with cold water by about an inch. Bring the water to a rolling boil over high heat. Once the water reaches a boil, immediately remove the pot from the heat and cover it. Let the eggs sit in the hot water for a period depending on your preference for doneness: 6 minutes for soft-boiled, 8 minutes for medium-boiled, and 12 minutes for hard-boiled eggs (as per Document 2). This method is consistent across Document 1 and Document 6, which also mention boiling eggs but focus more on the nutritional benefits and historical context, respectively.

  \end{mdframed}
\end{minipage} \label{fig:vd2}}%
\qquad
\subfloat[]{\begin{minipage}{\linewidth}
  \begin{mdframed}[roundcorner=10pt]
    To boil hard-boiled eggs, place the eggs in a single layer in a pot and cover with cold water by 1 inch. Bring to a boil over high heat, then remove from heat, cover, and let sit for 12 minutes for hard-boiled. For soft-boiled eggs, boil for 6 minutes, and for medium-boiled, 8 minutes. (Document 1)
  \end{mdframed}
\end{minipage} \label{fig:sd3}}%
\qquad
\subfloat[]{\begin{minipage}{\linewidth}
  \begin{mdframed}[roundcorner=10pt]
    To boil eggs, place them in a single layer in a pot and cover with cold water by about 1 inch. Bring the water to a boil over high heat. Once boiling, remove from heat, cover, and let sit in the hot water. For soft-boiled eggs, let them sit for 6 minutes; for medium-boiled, 8 minutes; and for hard-boiled, 12 minutes. This method is detailed in Document 1 (Document 1) and is supported by the basic boiling instructions found in Document 5 (Document 5) and Document 6 (Document 6), with Document 6 emphasizing the nutritional benefits of eggs for athletes and Document 5 highlighting the historical significance of boiling eggs.

  \end{mdframed}
\end{minipage} \label{fig:vd3}}%
\caption{(a) response generated using standard prompt in first pass, (b) response generated using VD prompt in first pass, (c) response generated using standard prompt in second pass, (d) response generated using VD prompt in second pass}%
\label{fig:ex2}%
 \end{figure}

\section{Experimental Setup} \label{expSetup}

We evaluated the Venn Diagram prompt on varied datasets from different fields. The aim of the prompt is to reduce the multi-step approach in long form, multiple document questions answering RAG systems. These datasets include questions from medical, financial, wikipedia articles, long context, long-form question answering datasets. The idea is to measure the efficacy of the prompt across datasets which require extensive text comparison, comprehensive context understanding and detailed narrative construction.

Models struggle particularly on long context documents \cite{reddy2024docfinqa}. Addressing these challenges may have a wide reaching impact across applications where specificity and long-range contexts are critical, like legal contract analysis, customer support issues and news articles to name a few.

\subsection{Dataset Description} \label{dd}

Existing work on QA datasets are often limited to short passages, preselected document sections or questions that can be answered directly with just one word or single sentence. It fails to reflect the broader and more realistic scenarios. The challenge faced by RAG systems with current datasets lies in effectively handling multi-hop queries in long-documents\cite{tang2024multihoprag}, which demands retrieving and reasoning over multiple pieces of supporting evidence. Therefore, we have selected datasets that specifically cover these challenges.

\begin{enumerate}
    \item We focused on documents which are complex, difficult to comprehend and characterized by lengthy contextual content. Eg: financial reports, ELI5.
    \item Secondly, we included a variety of context types ranging from single-doc/single-chunk to multi-doc/multi-chunk scenarios, ensuring that the RAG system's performance can be thoroughly assessed across different levels of complexity. 
    \item Lastly, we sought to evaluate prompt's capacity to establish inferences and engage in reasoning for different document types. We evaluated the prompt's comprehension capabilities to investigate patterns across diverse problem statements, which are influenced by the specific attributes of the documents.
\end{enumerate}

The overall intent is to select datasets that embody realistic industry scenarios and cover the above discussed challenges. Each dataset consists of a query, context and a ground truth answer.

Our method is evaluated on four benchmark datasets:

\begin{enumerate}
\item \textbf{ELI5}\cite{fan2019eli5}

ELI5 is one of the datasets from the KILT benchmark. The dataset is selected since it emphasizes the dual challenges of isolating relevant information within long source documents and generating multi-sentence answers from various sources in response to complex, diverse questions. There are 10 pre-generated samples available which have been used from the dataset.

\item \textbf{PubMedQA}\cite{jin2019pubmedqa}

PubMedQA represents a pioneering biomedical question answering (QA) dataset sourced from PubMed abstracts. It is chosen as it stands out as a well curated medical QA dataset demanding comprehensive reasoning over biomedical research texts, particularly emphasizing their quantitative aspects for answering questions. In this dataset, context length with 3000 characters or more is selected to test the model's comprehension and reasoning from detailed content.

\item \textbf{Long-Context QA}\cite{lcqa}

The dataset consists of minutes of government meetings as long context with a factual question. The ground truth is a short answer from the given context. Document context exceeding 15,000 characters length is selected to test the model's ability to generate accurate answers from lengthy and verbose documentation. 

\item \textbf{Sec 10-Q}\cite{sec10}

The collection comprises of financial documents tailored for advanced RAG applications across multiple documents. The dataset is developed in response to the gap in existing evaluation datasets, which are insufficient in representing the full spectrum of RAG use cases encountered in practical scenarios. Specifically, existing datasets predominantly focus on Q\&A over single or limited documents, whereas real-world scenarios often demand RAG over extensive document sets. 
\end{enumerate}

Across these four datasets, instructions in VD prompting maintains uniformity in all steps except the final one. The flexibility allows for customization as per on the desired output, such as generating a concise single-line answer or a more detailed response. It corresponds to the customization block as illustrated in Fig \ref{ragChanges}. These adjustments are made only to achieve the output in desired format without altering the underlying approach.

Various types of contexts covered across the four datasets are explained below :

\begin{enumerate}
    \item \textbf{Single-Document, Single-Chunk:} Queries where the answer resides within a contiguous segment (text or table chunk) of a single document. Successfully addressing these requires the prompt to extract the accurate chunk.
    \item \textbf{Single-Document, Multi-Chunk:} Queries where the answer spans multiple non-contiguous segments (text or table chunks) within a single document. Successfully addressing these necessitates the prompt to combine multiple relevant chunks from a single document, which could be challenging.
    \item \textbf{Multi-Document:} Queries where the answer spans multiple non-contiguous segments (text or table chunks) across multiple documents. Successfully addressing these requires the prompt to combine multiple relevant chunks from multiple documents.
\end{enumerate}

Table \ref{tab:avgChar} highlights the categorization of each of the datasets used along with insight into the average character length of contexts used.


\begin{table}[htbp]
\caption{Average Character length with Document Type}
\begin{center}
\begin{tabular}{|c|c|c|}
\hline
\textbf{Datasets} & \textbf{Document Type}& \textbf{Average} \\
& & \textbf{Context Length} \\
\hline
ELI5 & Single-Document, Multi-Chunk & 65235\\
\hline
PubMedQA & Single-Document, Multi-Chunk & 4500\\
\hline
Long-context QA & Single-Document, Single-Chunk & 18762.38\\
\hline
Sec 10-Q & Multi-Document & 650000\\
\hline
\end{tabular}
\label{tab:avgChar}
\end{center}
\end{table}

\subsection{Implementation Tools and Setup}

\subsubsection{LLM Model}

We use the state-of-the-art LLM GPT-4\cite{openai2024gpt4}. We selected this LLM because it has the strongest Chain-of-thought reasoning performance among public LLMs \cite{openai2024gpt4}. We ran all our experiments with 0.1 temperature setting.

\subsubsection{Prompts} \label{prompts}

To showcase the capabilities of the new prompt we compared a standard instruction to Venn Diagram approach based VD prompt. Some key aspects of both the prompt are:

\begin{itemize}
    \item \textit{Standard prompt}: All instructions in the prompt are presented with utmost clarity, precision, and without ambiguity, designed to facilitate optimal performance. The prompt is influenced from SEC 10-Q dataset achieving a 90\% efficacy rate with GPT-4.
    \item \textit{VD prompt}: The prompt models the Venn Diagram approach. It is instructed to extract the relevant information with respect to the query and organize the distinct and overlapping information similar to Venn Diagram approach. It helps the prompt to provide a consolidated view, recognize patterns and trends, and facilitate comparisons and enhance understanding. 
\end{itemize}



\subsubsection{Evaluation Criteria} \label{evalcriteria}

We employed two evaluation mechanisms: the RAGAS\cite{es2023ragas} metric evaluation framework and LLM-as-a-judge model\cite{zheng2023judging}. The details of each of the evaluation mechanisms are captured in the following sub-sections \ref{ragas} and \ref{llmJudge}.

\paragraph{\textbf{RAGAS}} \label{ragas}

    \begin{itemize}
        \item The framework provides quantifiable metrics to assess RAG pipeline performance across retrieval and generation stages. We utilized the metrics to evaluate only the generation module.
        \item The metrics for evaluating generation module include:
        \begin{itemize}
            \item Answer Relevance: Evaluates whether the generated text addresses the query or meets the user's purpose, serving as a metric for the overall efficacy of the RAG output.
                \begin{equation}
                AnswerRelevance = \dfrac{1}{N}\sum_{i=1}^{N}cos(E_{g_i}, E_o)
                \end{equation}
                $E_{g_i}$: Embedding of generated question i
                \newline
                $E_o$: Embedding of original question
                \newline
                $N$: Number of generated questions
            \item Answer Semantic Similarity: Refers to the evaluation of how closely the generated answer aligns with the semantic content of the ground truth.
                \begin{equation}
                AnswerSimilarity = cos(E_{gt}, E_a)
            \end{equation}
            $E_{gt}$: Embedding of ground truth answer
            \newline
            $E_a$: Embedding of generated answer
            \item Answer Correctness: Involves assessing the accuracy of the generated answer in comparison to the ground truth.
                \begin{equation}
                AnswerCorrectness = \dfrac{
                \begin{multlined}[0.1\linewidth]
                (W_1 * AnswerSimilarity) \\ + (W_2 * F1-Score)
                \end{multlined}
                }{W_1 + W_2}  
            \end{equation}
            $W_1$, $W_2$: Weights assigned to each metric 
        \end{itemize}
    \end{itemize}
    
\paragraph{\textbf{LLM-as-a-Judge Model}} \label{llmJudge}

\begin{itemize}
    \item It is a promising approach utilizing LLMs themselves as judges \cite{openai2024gpt4}.  We leverage an LLM to assess outputs in a human-like manner, reducing the requirement for human intervention.
    \item It overcomes the challenges and complexity of devising a rule-based program and traditional metrics like ROUGE\cite{lin-2004-rouge} or BLEU\cite{10.3115/1073083.1073135}, as they are often inadequate for nuanced tasks such as answer evaluation.
\end{itemize}

We have used two different prompts for LLM-as-a-judge model to cover different aspects. It is done to overcome the limitations of LLM-as-a-judge which suffers from position bias and inconsistent results\cite{wang2023large}. The approach helped to establish consistent and more accurate evaluation scores.

\begin{enumerate}
    \item LLM Judge-1 is a reference-free evaluator that inputs Question Q, Knowledge K, Answer A into the evaluation prompt. We adopted the multi-dimensional evaluation strategy to gauge answer relevancy and correctness. We incorporated dimensions such as explicitness, helpfulness, directness, grammaticality, relevance, edge-case reasoning, factuality, supposition, objectivity, creativity, hallucination and source credibility. This is motivated by existing prompt evaluation template\cite{llmjudge1}.
    \item LLM judge-2 functions as a text comparison evaluator, examining solely the Ground Truth Answer (GT) and the newly generated Answer (A). It assesses Answer A in comparison to Ground Truth Answer GT, considering factors such as relevance, completeness, semantic coherence, and lexical similarity. This evaluation aims to ensure that Answer A aligns closely with the expected standards established by Ground Truth Answer GT.
\end{enumerate}

\section{Evaluation Results} \label{evaluationResults}

For each of the datasets described in Section \ref{dd}, we generate two sets of responses. The first response is generated from the proposed VD prompt and the second response is generated using the instruction based standard prompt following best practices and conforming to optimal guidelines. The exact details pertaining to each of these prompts are provided in Section \ref{prompts}. The responses from each of these prompts are evaluated against the query, reference context, and a ground truth response already provided as part of the dataset. 

\subsection{Results from RAGAS}

Table \ref{tab:ragasScores} highlights the metric scores achieved by using RAGAS and obtaining average values over all samples in each of the datasets used for experimentation. RAGAS evaluation predominantly focuses on three sets of metrics: answer relevancy, answer similarity, and answer correctness.

\begin{table}[htbp]
\caption{RAGAS Metric Scores}
\begin{center}
\begin{tabular}{|c|c|c|c|}
\hline
 & \textbf{Answer}& \textbf{Answer} & \textbf{Answer} \\
 & \textbf{Relevancy} & \textbf{Similarity} & \textbf{Correctness} \\
\hline
\textbf{PubMedQA} & & & \\
\hline
VD Prompt & 0.9597 & 0.9172 & 0.5353 \\
\hline
Standard Prompt & 0.9427 & 0.8690 & 0.3838 \\
\hline
\textbf{ELI5} & & & \\
\hline
VD Prompt & 0.9027 & 0.9026 & 0.4466 \\
\hline
Standard Prompt & 0.9056 & 0.8904 & 0.3880 \\
\hline
\textbf{Long Context QA} & & & \\
\hline
VD Prompt & 0.9549 & 0.9118 & 0.8017 \\
\hline
Standard Prompt & 0.9465 & 0.9083 & 0.7517 \\
\hline
\textbf{Sec 10-Q} & & & \\
\hline
VD Prompt & 0.9353 & 0.9471 & 0.5512 \\
\hline
Standard Prompt & 0.9273 & 0.9575 & 0.6951 \\
\hline
\end{tabular}
\label{tab:ragasScores}
\end{center}
\end{table}

\begin{figure}%
\centering
\subfloat[]{{\includegraphics[width=8cm]{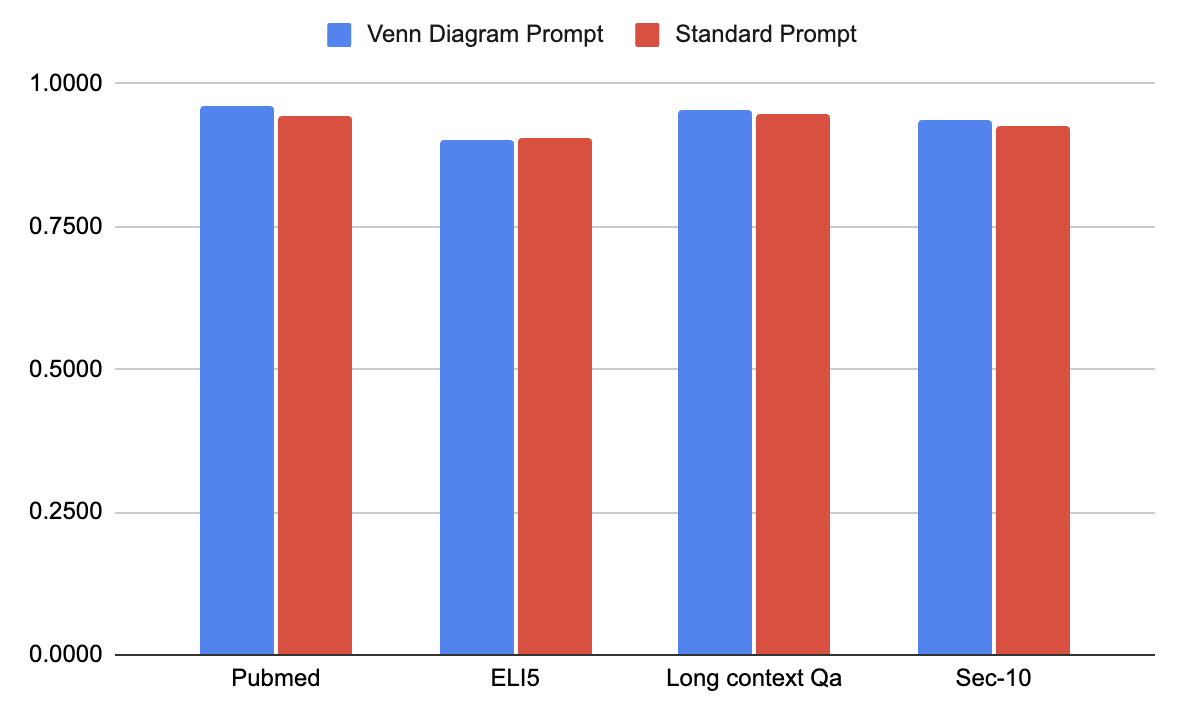} \label{fig:ar} }}%
\qquad
\subfloat[]{{\includegraphics[width=8cm]{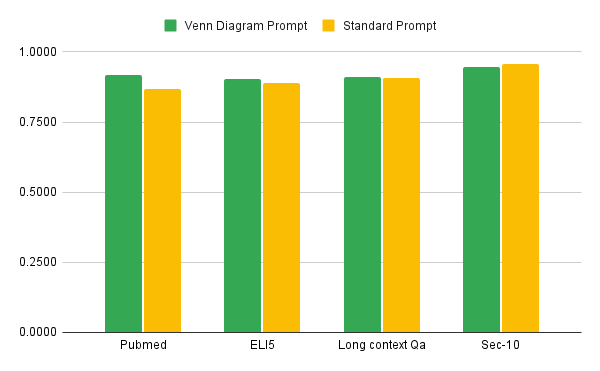} } \label{fig:as}}%
\qquad
\subfloat[]{{\includegraphics[width=8cm]{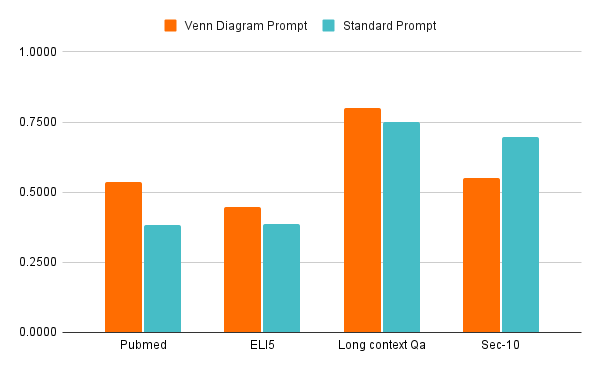} } \label{fig:ac}}%
\caption{(a) Answer Relevance Score Comparison, (b) Answer Semantic Similarity Score Comparison, (c) Answer Correctness Score Comparison}%
\label{fig:ragasGraphs}%
 \end{figure}

Answer Relevancy concentrates on evaluating the extent to which the provided answer aligns with the given query. The criterion for Answer Relevancy is to assign a score based on the average of the cosine similarity between the original question and the additional artificial questions that can be created from the generated answer by reverse engineering. Fig. \ref{fig:ar} showcases the Answer Relevancy scores across all the four datasets used, by comparing the proposed VD prompt against the instruction-based standard prompt. For almost all the datasets, VD prompt provides a competitive score across all the datasets. Thus, answer relevancy highlights that by focusing specifically on the answer generated and the original question, we are able to achieve higher or comparative scores with VD prompt than the standard prompting technique.

Answer Semantic Similarity, as the name suggests, is predominantly concerned with the semantic similarity of the generated response in comparison to the ground truth answer provided as part of the dataset. The comparison between the two responses is measured by calculating the cosine similarity of the vector representation of the answers. Fig. \ref{fig:as} shows that VD prompt provides a consistent improvement with respect to almost all the datasets except Sec 10-Q. However, it is important to note that we use the same prompt for our standard prompt as the one provided by the authors of Sec 10-Q which was used for generating their ground truth responses. As a result, an inherent bias is introduced during answer similarity calculation as represented in our analysis as well.

Answer Correctness evaluates the generated answer similarity aspect in comparison to the ground truth response while also encapsulating the factual correctness. The idea of the metric score is to provide an overall measure of the accuracy of the generated response. With respect to this specific metric, we observe from Fig. \ref{fig:ac} that VD prompt provides a drastic improvement in comparison to the standard prompt for almost all the datasets except Sec 10-Q. As explained in Answer Semantic Similarly comparison, the deviation in this metric for Sec 10-Q can be attributed to the fact that the same standard prompt is used for both answer generation and ground truth generation.

\subsection{Results from LLM-as-a-Judge Model}
The score for a generated answer represents the cumulative sum of the LLM-as-a-Judge model's evaluation score on the above mentioned dimensions, where each dimension can get a score of either 0 or 1. It is applied for both of the LLM Judge approaches.
LLM Judge 1 provides a score ranging from 0 to 10, since it has 10 dimensions, measuring relevance of the generated answers from each of the prompts with respect to the user query and contexts provided.
Similarly, LLM Judge 2 scores range from 0 to 5 by comparing generated answer from each of the prompts with the ground truth.
Table \ref{tab:llmScores} shows evaluation results by using LLM-as-a-judge model for the two approaches. 
For each prompting technique, the average score is calculated across all the samples in a dataset for both LLM Judges.
To keep the evaluation fair, we ensured that neither of the LLM Judges were aware of which prompt the generated answer corresponded to.

\begin{table}[htbp]
\caption{LLM-as-a-Judge Metric Scores}
\begin{center}
\begin{tabular}{|c|c|c|}
\hline
 & \textbf{LLM Judge 1}& \textbf{LLM Judge 2} \\
\hline
\textbf{PubMedQA} & & \\
\hline
VD Prompt & 7.8182 & 3.7273 \\
\hline
Standard Prompt & 6.7273 & 2.4545 \\
\hline
\textbf{ELI5} & & \\
\hline
VD Prompt & 7.454 & 3.1818 \\
\hline
Standard Prompt & 6.9000 & 2.7000 \\
\hline
\textbf{Long Context QA} & & \\
\hline
VD Prompt & 8.4545 & 3.7273 \\
\hline
Standard Prompt & 7.7273 & 3.1818 \\
\hline
\textbf{Sec 10-Q} & & \\
\hline
VD Prompt & 8.4737 & 3.4737 \\
\hline
Standard Prompt & 8.1000 & 3.7000 \\
\hline
\end{tabular}
\label{tab:llmScores}
\end{center}
\end{table}

\begin{figure}%
\centering
\subfloat[]{{\includegraphics[width=8cm]{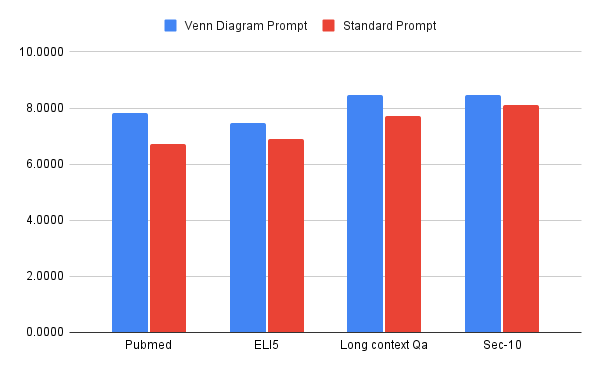} \label{fig:llm1} }}%
\qquad
\subfloat[]{{\includegraphics[width=8cm]{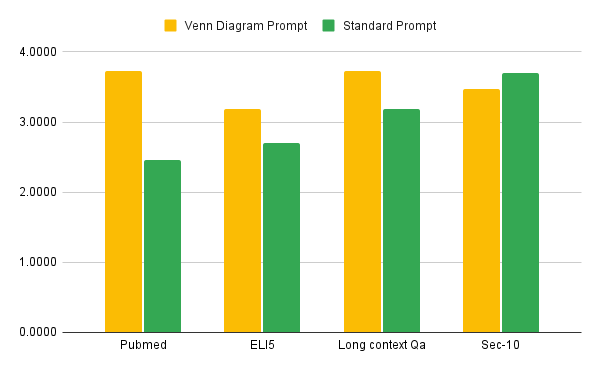} } \label{fig:llm2}}%
\caption{(a) Score Comparison from LLM-as-a-judge model 1, (b) Score Comparison from LLM-as-a-judge model 2}%
\label{fig:ragasGraphs}%
 \end{figure}

Fig. \ref{fig:llm1} compares the results from the LLM Judge 1 which focuses on providing an evaluation score on relevance of the generated answer when compared with the query and context document(s). As per the evaluation, VD prompt consistently outperforms standard prompt across all the datasets. The dimensions defined in the LLM Judge 1 found the responses from VD prompt to be more consistent with the query and devoid of hallucination when compared to the relevant context.

Results from LLM Judge 2 are shown in Fig. \ref{fig:llm2} where ground truth response was compared with the generated response. LLM Judge 2 found results from the VD prompt to be more precise and similar to the ground truth response in comparison to the standard prompt across all datasets except Sec 10-Q. The deviation for Sec 10-Q is in-line with the similar comparison results obtained from RAGAS.

In conclusion, based on the evaluation results, that span from Long context QA to more complicated tasks such as PubMed or Sec 10-Q, VD prompt consistently performs better than a meticulously crafted instruction prompt which adheres to optimal guidelines and practices. It is also worth noting that although Answer Semantic Similarity and Answer Correctness of Sec 10-Q is lower for VD prompt, a higher score for Answer Relevancy highlights that the response from VD prompt is potentially more relevant to the query, which is also validated by LLM-as-a-Judge model's results. 

Additionally, the analysis highlights that VD prompt provides improvement in results for long form answer generation irrespective of the domain of the dataset. This is in addition to all the significant benefits that VD prompt inherently provides as described in Section \ref{adv}. 

The success of VD prompt is attributed to the prompt's structured organization ability, enabling it to provide consistently accurate responses with respect to the query and context, which is difficult to achieve with a standard prompt.

\section{Advantages} \label{adv}

\begin{itemize}
    \item \textbf{Fact-Checking}: The VD prompt provides a detailed response clearly citing the source of each information present with relevance to the answer, instead of a response with consolidated source list without their relevance to the answer. These explicit citations act as a fact checking tool enabling a human evaluator to quickly validate whether the information is present in the cited source or not, which helps in checking for hallucinations.

    \item \textbf{Rich references for Information Retrieval}: Due to the nature of the VD prompting technique, each detail is supported by its corresponding source document, and the cited sources are presented with the relevance to the answer. It results in a richer and more refined retrieval since it highlights the appropriate sections of the source document(s) that are referred in the final response. This feature is difficult to implement with a standard prompt which provides citations on a broader scale. The existing LLM based Information Retrieval and AI Chatbots also suffer from this limitation of being unable to provide sources with relevance to the generated answer.
\end{itemize}

\section{Future Work} \label{futureScope}

While the VD prompt has demonstrated competitive and outstanding results in comprehending complex documents, there exist avenues for further enhancement. 
We propose investigating the utility of VD prompting as a re-ranker mechanism to select the most relevant documents. Re-ranking involves reordering the document in the post-retrieval stage to further refine relevance and accuracy of the final answer. This paper lays the theoretical foundations of leveraging the prompt to find document relevance. Furthermore, the model demonstrates enhanced understanding and reasoning capabilities. This is credited to the initial organization and categorization of information which play a crucial role in subsequent reasoning processes. The model reflects on the evidence found before formulating a final response. Therefore, the VD prompt's inherent ability to organize, analyze, and reflect on data presents the potential to excel in complex analytical tasks such as identifying trends across financial reports and appears to offer a credible path forward. 

Through this exploration, we aim to elucidate the potential of a Venn diagram based prompting technique to highlight avenues for future research in this domain.

\section{Conclusion} \label{conclusion}
We have introduced a novel prompting technique, VD prompt. We explain the intuition of this approach using set theory and illustrate it with an example. We have conducted empirical research demonstrating the efficacy of VD prompting in producing accurate answers from lengthy contextual documents. Our approach involves utilizing varied and realistic question answering datasets, coupled with thorough evaluations employing RAGAS and LLM-as-a-judge model. VD prompt consistently achieved higher or comparative scores as compared to the standard prompting technique. 

Finally, we conclude, this innovative workflow enhances the generation module within RAG systems. Moreover, it effectively tackles the ``Lost in the Middle" issue encountered by LLMs when dealing with extensive context. This method organizes information in a structured manner, providing linguistic feedback to the model. Consequently, the model becomes less reliant and sensitive on the positioning of information in the provided context.

\bibliographystyle{plain}
\bibliography{VDP.bib} 

\end{document}